\documentclass{article}

     \PassOptionsToPackage{numbers, compress}{natbib}

     \usepackage[preprint]{neurips_2019}

\usepackage[utf8]{inputenc} 
\usepackage[T1]{fontenc}    
\usepackage{hyperref}       
\usepackage{url}            
\usepackage{booktabs}       
\usepackage{amsfonts}       
\usepackage{nicefrac}       
\usepackage{microtype}      
\usepackage{graphicx}
\usepackage{csquotes}
\usepackage{subcaption}
\usepackage{makecell}
\usepackage{comment}
\usepackage{amsmath}
\usepackage{import}
\usepackage{tabulary}
\usepackage{float}
\usepackage{caption}
\usepackage{wrapfig}
\usepackage{lipsum}
\usepackage{float}

\title{Individual predictions matter: Assessing the effect of data ordering in training fine-tuned CNNs for medical imaging}

\author{%
  John R. Zech\thanks{equal contribution} \\
  Department of Radiology\\
  Columbia University Irving Medical Center\\
  New York, NY \\
  \texttt{jrz2111@columbia.edu} \\
   \And
   Jessica Zosa Forde$^*$, Michael L. Littman \\
   Department of Computer Science \\
   Brown University \\
   Providence, RI \\
   \texttt{jessica\_forde@brown.edu} \\ \texttt{mlittman@cs.brown.edu}}

\begin{document}

\maketitle

\begin{abstract}
   We reproduced the results of CheXNet with fixed hyperparameters and 50 different random seeds to identify 14 finding in chest radiographs (x-rays). Because CheXNet fine-tunes a pre-trained DenseNet,
   the random seed affects the ordering of the batches of training data but not the initialized model weights. We found substantial variability in predictions for the same radiograph across model runs (mean $\ln(P_{\text{max}}/P_{\text{min}})$ 2.45, coefficient of variation 0.543). This individual radiograph-level variability was not fully reflected in the variability of AUC on a large test set.
   Averaging predictions from 10 models reduced variability by nearly 70\% (mean coefficient of variation from 0.543 to 0.169, t-test 15.96, p-value < 0.0001). We encourage researchers to be aware of the potential variability of CNNs and ensemble predictions from multiple models to minimize the effect this variability may have on the care of individual patients when these models are deployed clinically. 

\end{abstract}

\section{Introduction}

While there is interest in using convolutional neural networks (CNNs) to identify findings in medical imaging \citep{Rajpurkar2017-nh, Irvin2019-lc,Johnson2019-as,Wang2017-py,Wang2018-ry,Zech2018-tg}, some researchers have questioned the reliability \citep{Sculley2015-da, Papernot2016-lz, Rahimi2017-xb, Finlayson2019-ed, Forde2019-fv} and reproducibility \citep{Vaswani2017-iu, Melis2018-mv, Lucic2018-bp, Riquelme2018-cn, Henderson2018-yx} of deep learning methods. Often, researchers in medical imaging evaluate a single model's predictions to measure performance \citep{gale,mabal,Rajpurkar2017-nh,tienet}. However, the loss surface of a CNN is non-convex, and differences in training such as random seed \citep{Henderson2018-yx} and optimization method \citep{Wilson2017-ca, Choi2019-od} can affect the learned model weights and, consequently, the predictions of the model. Within the machine learning community, there are concerns that evaluation of a single trained model does not provide sufficient measurement of its variability \citep{Henderson2018-yx, Forde2019-fv}.  To mitigate this variability, some researchers in medical imaging have used cross-validation \citep{thrun, Chang2018-ch}, which varies the data used for both training and testing, but still uses a single model to make predictions during each cross-validation run. Others have used ensembling \citep{Sollich1996-wj}, combining predictions from 3 \citep{titano}, 5 \citep{wu,mura}, 10 \citep{Gulshan2016-mb, Rajpurkar2018-vn, Pan2019-ux}, and 30 \citep{Irvin2019-lc} different trained models to optimize classification performance. While ensembles have been proposed as a simple method for estimating the uncertainty of the predictions of a CNN \citep{Lakshminarayanan2017-en, Ovadia2019-qr}, such variability measurements, as seen in \citet{wu}, are not common in medical imaging research. 

If models generate different predictions when they are retrained, they may make inconsistent predictions for the same patient. The AUC of a single trained model will not give a direct indication of this inconsistency. AUC itself will likely vary between retrained models, which could complicate efforts to compare CNN performance to other models or human experts using statistical testing \citep{Silva-Aycaguer2010-fu}. Researchers have attempted to quantify the uncertainty of predictions for individual patients by statistical estimation \citep{Schulam2019-ui} and direct prediction \citep{Raghu2019-uk}. \citet{Raghu2019-uk} proposed a machine learning method which identifies which retinal fundus photographs \citep{Gulshan2016-mb} would be likely to have human expert disagreement in diagnosing diabetic retinopathy; such a model could be used to identify high uncertainty cases likely to benefit from a second opinion. \citet{Dusenberry2019-xh} examined the variability of RNN-based mortality prediction using the medical records of ICU patients in MIMIC-III~\citep{Johnson2019-as} and recommended the use of Bayesian RNNs \citep{Fortunato2017-fo} with stochastic embedding over ensembling as a way to estimate the variability of predictions in clinical time-series data.  

In this study, we explicitly characterized the variability in individual predicted findings and overall AUC of a CNN that was trained multiple times to predict findings on chest radiographs.  Like \citet{Dusenberry2019-xh}, we found notable variability of predictions on individual patients with similar aggregate performance metrics. 
Because many real-world 
clinical decision support systems rely on single values of predicted probability rather than statistical distributions incorporating uncertainty, we focused our analysis on the use of ensembling for the purposes of robust prediction.  We found that, in the case of chest radiographs, simple ensembling can reduce the variability of these probability estimates; ensembles of as few as ten models were found to reduce the variability of predictions by 70\% (mean coefficient of variation from 0.543 to 0.169, t-test 15.96, p-value < 0.0001).  

\section{Methods}

Using an open source implementation \citep{Zech2018-gf}, we replicated the model described in \citet{Rajpurkar2017-nh} 50 times, varying the random seed with each fine-tuning \citep{Lakshminarayanan2017-en}.  Per \citet{Rajpurkar2017-nh}, a DenseNet-121 \citep{Huang2017-hi} pre-trained on ImageNet \citep{Russakovsky2015-ib} was fine-tuned
to identify 14 findings in the NIH chest radiography dataset ($n$=112,120) \citep{Wang2017-py}. 
The dataset was partitioned into 70\% train, 10\% tune, and 20\% test data (train $n$=78,468, tune $n$=11,219, test $n$=22,433). These 50 models were fine-tuned using SGD with identical hyperparameters on the same train and tune datasets. Because the CNN was consistently initialized with parameters from a DenseNet-121 pre-trained to ImageNet, the only difference in the training procedure across model runs was the order in which data was batched and presented to the model during each epoch of fine-tuning. Each model’s performance was assessed on the full test partition ($n$=22,433). To replicate the test set used for labeling by radiologists \citep{Rajpurkar2017-nh}, a smaller test partition ($n$=792) was created by randomly sampling 100 normal radiographs and 50 positive examples for each finding except for hernia ($n$=42 in the test set).

\begin{table}[hbp]
\caption{Variability statistics for each finding on a radiograph in the test set across all 50 models. To measure the relative magnitude of predictions, we calculated the percentile rank of a prediction relative to all predictions for that finding in the test set and report the percentile rank range.}
\label{table-metrics}
\begin{tabulary}{\textwidth}{LCL}
\hline 
\textbf{Metric}& \textbf{Symbol} & \textbf{Description (for each finding, radiograph pair)} \tabularnewline
\hline 
Mean & $\mu$ & Average predicted probability\tabularnewline
\hline 
Stdev. & $\sigma$ & Standard deviation of predicted probability\tabularnewline
\hline 
Coefficient of variation & $\sigma/\mu$ & Standard deviation divided by mean predicted probability\tabularnewline
\hline 
Log prob. ratio & $\ln(\frac{P_{\text{max}}}{P_{\text{min}}})$ & Natural log of the highest predicted probability divided by the lowest
predicted probabilty\tabularnewline
\hline 
Percentile (\%ile) rank & $R(P)$ & The percentile rank relative to all predictions for that finding among radiographs in the test set\tabularnewline
\hline 
\%ile rank range & $R(P_{\text{max}})-R(P_{\text{min}})$  & The percentile rank of the highest probability predicted minus the percentile
rank of the lowest probability predicted   \tabularnewline
\hline 
\end{tabulary}
\end{table}

We calculated various statistical measurements for each finding on each radiograph across our 50 models in the full test set ($n$=22,433). Table \ref{table-metrics} describes each metric in detail. In addition to mean, standard deviation, and coefficient of variation, we calculated $\ln(\frac{P_{\text{max}}}{P_{\text{min}}})$,
where $P_{\text{max}}$ is the greatest and $P_{\text{min}}$ the least probability predicted by the 50 trained models for a given finding on a given radiograph. This ratio provides a scaled measurement of the variability of the predicted probability of a finding on a radiograph. To contextualize predictions within the population of predictions for that finding, we calculated the percentile rank range of the predictions, $R(P_{\text{max}}) - R(P_{\text{min}})$, where $R(P)$ is the percentile rank of a prediction relative to all predictions for that finding in the test set. 

To evaluate the effectiveness of ensembling in reducing variance, we averaged predictions over disjoint groups of 10 models to yield 5 separate averaged predictions for each finding for each radiograph ($n$=5 groups $\times$ 10 models per group = 50 total models). We reported the standard deviation and coefficient of variation across these averaged groups. A paired t-test was used to compare coefficients of variation across the raw ($n$=50) and averaged ($n$=5) predictions. 

We examined the variance in overall AUC for the 14 possible targets. For each of the 50 models, AUC was calculated on both the full and limited test sets using the pROC package in R \citep{Robin2011-lg,Ihaka1996-kl}. This calculation provided an empirical distribution of the test AUC relative to the order of samples in the training data. For both the full and limited test sets, we calculated the 95\% confidence interval of this distribution by subtracting the second smallest AUC from the second largest of the 50 AUCs. For the limited test sets, the average width of the 95\% confidence interval was also estimated using DeLong’s method~\citep{DeLong1988-je} and bootstrapping \cite{Carpenter2000-xn}. DeLong
expresses AUC in terms of the Mann-Whitney U statistic \citep{Mann1947-kd}, a non-parametric test statistic that is approximately normally distributed for large sample size, and thus can used to calculate confidence intervals.

\section{Results}

\begin{figure*}[h]
  \centering
  \includegraphics[width=0.6\textwidth]{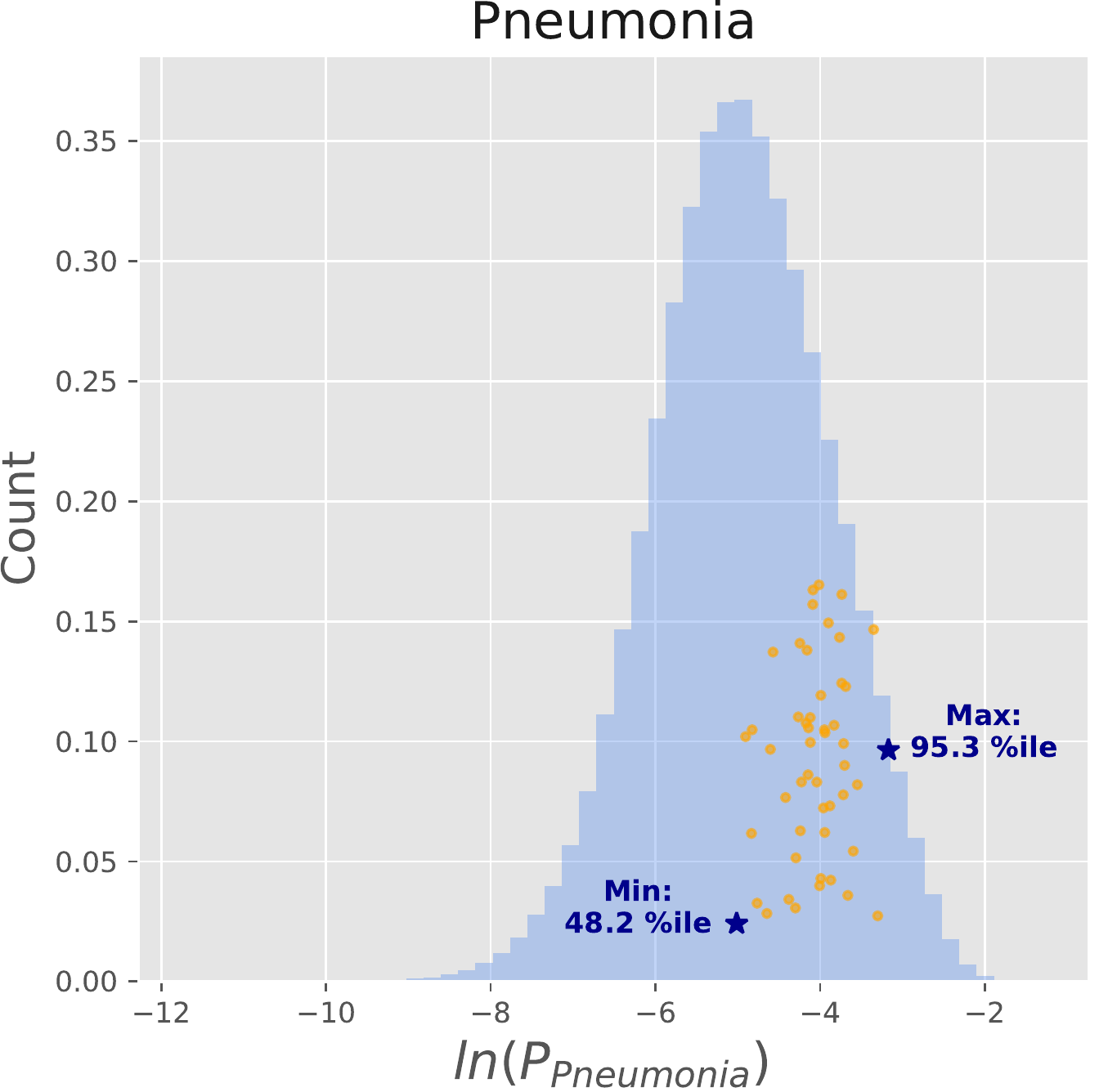}
  \caption{Comparison of the variability in pneumonia prediction made for a single radiograph across trained models (jittered orange dots and blue stars, $n$=50) to the variability of the full test set across trained models (blue histogram, $n$=22,433 cases $\times$ 50 models = 1,121,650). The predicted risk of pneumonia on the single radiograph ranged from the 48.2 to the 95.3 percentile.}
  \label{fig:lnp_scatter_pna_only}
\end{figure*}

The variability in predictions for a given radiograph was substantial across models. An example radiograph classification is shown in Figure~\ref{fig:lnp_scatter_pna_only}. Figure~\ref{fig:lnp_scatter_pna_only} compares the variability across models ($n$=50) in predicted probability of pneumonia for this radiograph to the variability of predictions for pneumonia in the full test set ($n$=22,433 cases $\times$ 50 models = 1,121,650 total predictions). In this example, the percentile rank range was 95.3\% $-$ 48.2\% = 47.1\%. The variability of each finding for this radiograph relative to all predictions for a given finding is shown in Figure~\ref{fig:lnp_scatter}. 

\begin{table}[htbp]
\caption{Average variability metrics for individual models and ensembles. Note that coefficient of variation, $\sigma/\mu$, is mean($\sigma/\mu$), not mean($\sigma$)/mean($\mu$).  Varying the order of the training data resulted in varying predictions for the same radiograph. In the full test set ($n$=22,433), a single radiograph had an average $\ln(P_{\text{max}}/P_{\text{min}})$ of 2.45, coefficient of variation of 0.543, and percentile rank range, $R(P_{\text{max}}) - R(P_{\text{min}})$, of 43.0\%. Averaging across ten models significantly reduced this variability.
}
\label{table-pmax-pmin}
\centering
\addtolength{\tabcolsep}{-1pt} 
\begin{tabular}{cccccccc}
\hline
                            & \multicolumn{5}{c}{\textbf{Average across individual models (n=50)}}                                                        & \multicolumn{2}{c}{\begin{tabular}{@{}c@{}}\textbf{Ensemble of} \\ \textbf{ 10 runs (n=5)}\end{tabular}}  \\
\hline
\textbf{Finding}          & \textbf{Mean ($\mathbf{\mu}$)} & \textbf{Stdev. ($\sigma$)} & \begin{tabular}{@{}c@{}}\textbf{$\sigma/\mu$}\end{tabular}  &
\begin{tabular}{@{}c@{}}$\mathbf{\ln(\frac{P_{max}}{P_{min}}})$ \end{tabular} &
\begin{tabular}{@{}c@{}}\textbf{\%ile rank}\\\textbf{range}\end{tabular} &
\textbf{$\sigma$} & 
\begin{tabular}{@{}c@{}}\textbf{$\sigma/\mu$} \end{tabular}                       \\
\hline
\textbf{Atelectasis}        & 0.107         & 0.034         & 0.449                   & 2.085                              & 0.360                   & 0.011         & 0.142                                               \\
\textbf{Cardiomegaly}       & 0.030          & 0.014         & 0.686                   & 2.993                              & 0.404                  & 0.004         & 0.211                                               \\
\textbf{Consolidation}      & 0.041         & 0.014         & 0.439                   & 2.046                              & 0.368                  & 0.004         & 0.133                                               \\
\textbf{Edema}              & 0.022         & 0.009         & 0.654                   & 2.921                              & 0.378                  & 0.003         & 0.205                                               \\
\textbf{Effusion}           & 0.128         & 0.033         & 0.523                   & 2.415                              & 0.309                  & 0.010          & 0.163                                               \\
\textbf{Emphysema}          & 0.023         & 0.010          & 0.703                   & 3.033                              & 0.479                  & 0.003         & 0.219                                               \\
\textbf{Nodule}             & 0.056         & 0.021         & 0.444                   & 2.029                              & 0.493                  & 0.007         & 0.140                                                \\
\textbf{Pneumonia}          & 0.012         & 0.004         & 0.403                   & 1.867                              & 0.451                  & 0.001         & 0.126                                               \\
\textbf{Fibrosis}           & 0.016         & 0.007         & 0.531                   & 2.435                              & 0.446                  & 0.002         & 0.171                                               \\
\textbf{Hernia}             & 0.002        & 0.001        & 0.608                  & 2.784                              & 0.494                  & 0.0004        & 0.185                                               \\
\textbf{Infiltration}       & 0.172         & 0.042         & 0.299                   & 1.401                              & 0.425                  & 0.013         & 0.091                                               \\
\textbf{Mass}               & 0.051         & 0.022         & 0.624                   & 2.765                              & 0.493                  & 0.007         & 0.199                                               \\
\textbf{Pleural Thickening} & 0.029         & 0.012         & 0.515                   & 2.367                              & 0.457                  & 0.004         & 0.162                                               \\
\textbf{Pneumothorax}       & 0.046         & 0.022         & 0.723                   & 3.196                              & 0.465                  & 0.007         & 0.227                             \\
\hline


\end{tabular}
\end{table}

\begin{figure*}[h]
  \centering
  \includegraphics[width=1.0\textwidth]{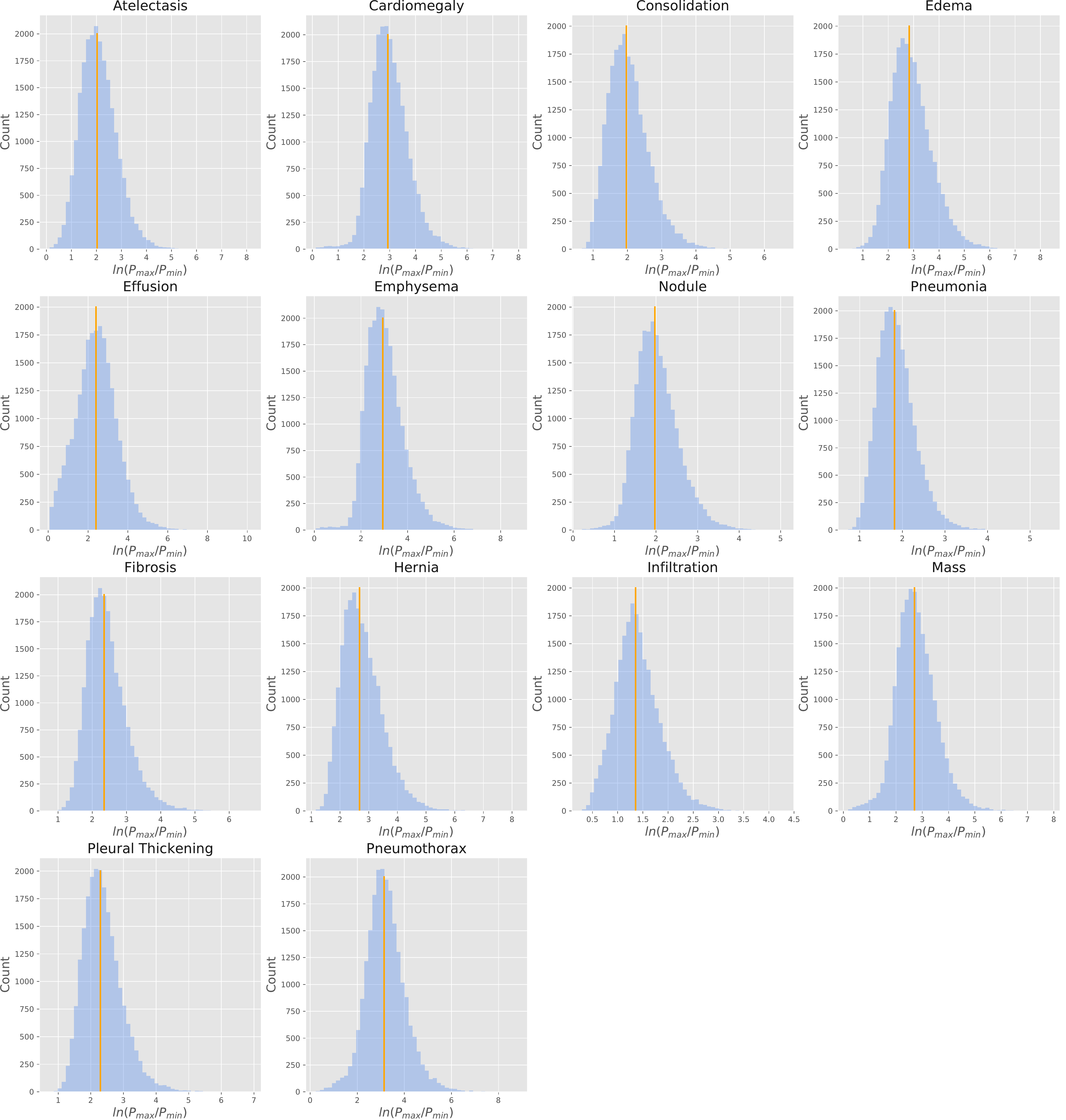}
  \caption{Distribution of $\ln(\frac{P_{\text{max}}}{P_{\text{min}}})$ for each finding in the full test set ($n$=22,433), where where $P_{\text{max}}$ is the highest and $P_{\text{min}}$ the lowest probability predicted for a given finding on a given radiograph across all 50 models. The mean $\ln(\frac{P_{\text{max}}}{P_{\text{min}}})$ was 2.45, indicating substantial variability in predictions made for the same radiograph; models with consistent predictions would have mean $\ln(\frac{P_{\text{max}}}{P_{\text{min}}})$ of zero.}
  \label{fig:ln-pmax-pmin}

\end{figure*}

In the full test set ($n$=22,433),
the mean coefficient of variation for an individual radiograph over 50 retrainings was 0.543, and mean $\ln(\frac{P_{\text{max}}}{P_{\text{min}}})$ was 2.45 (Table~\ref{table-pmax-pmin}, Figure~\ref{fig:ln-pmax-pmin}); 
for a model with unvarying predictions, $\ln(\frac{P_{\text{max}}}{P_{\text{min}}})$ would equal zero. The radiographs had a mean percentile rank range of 43.0\%. In other words, the average difference between the percentile rank of a radiograph's highest prediction, relative to all predictions for that finding in the test set, and the radiograph's lowest prediction of that finding, was 43.0\%---nearly half the available range. 

\begin{figure}[h] 
\centering

\begin{subfigure}[t]{.8\textwidth}
\centering
\includegraphics[width=\linewidth]{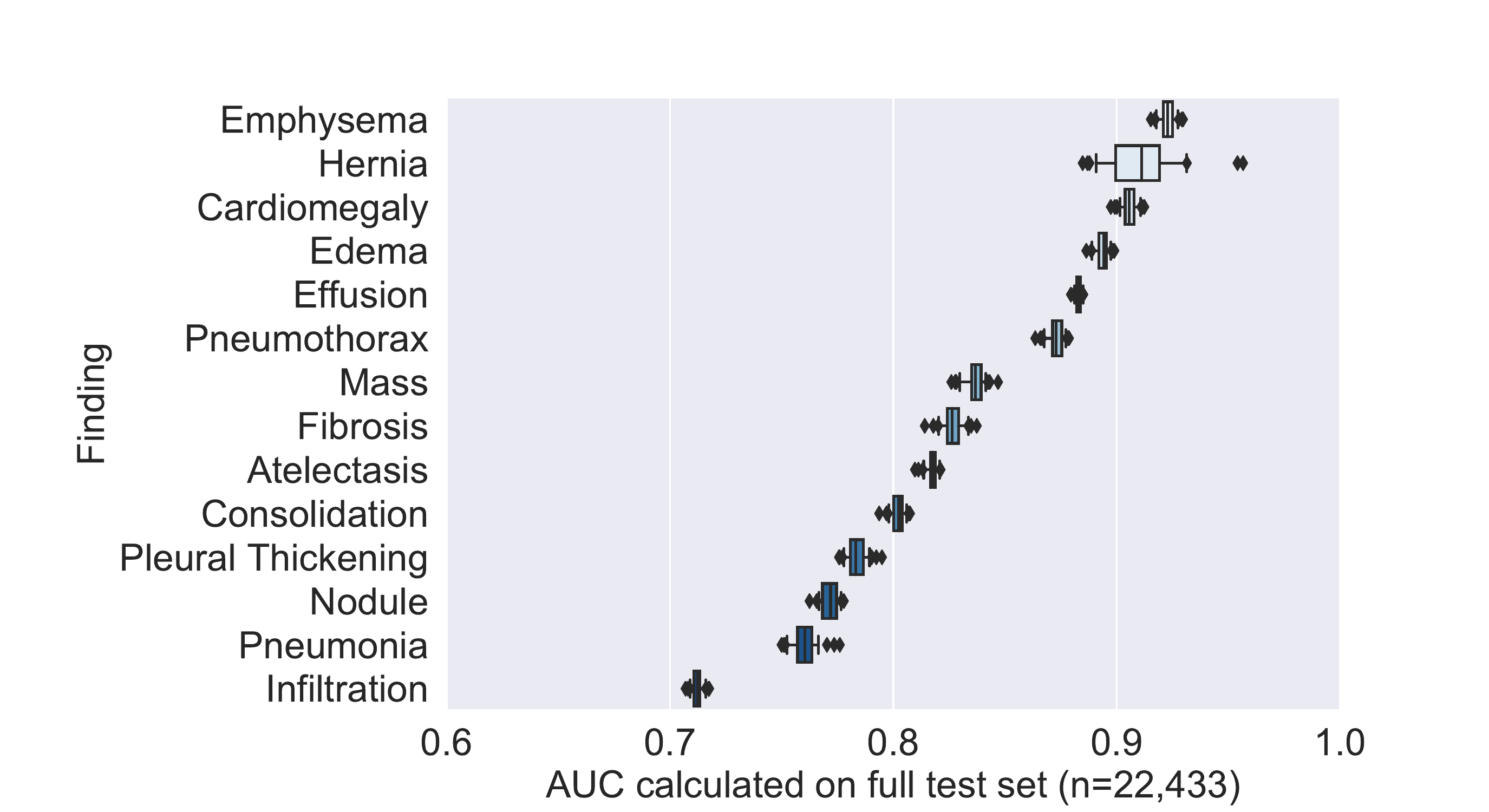}
        \label{fig:fig_a}
\end{subfigure}
\begin{subfigure}[t]{.8\textwidth}
\centering
\includegraphics[width=\linewidth]{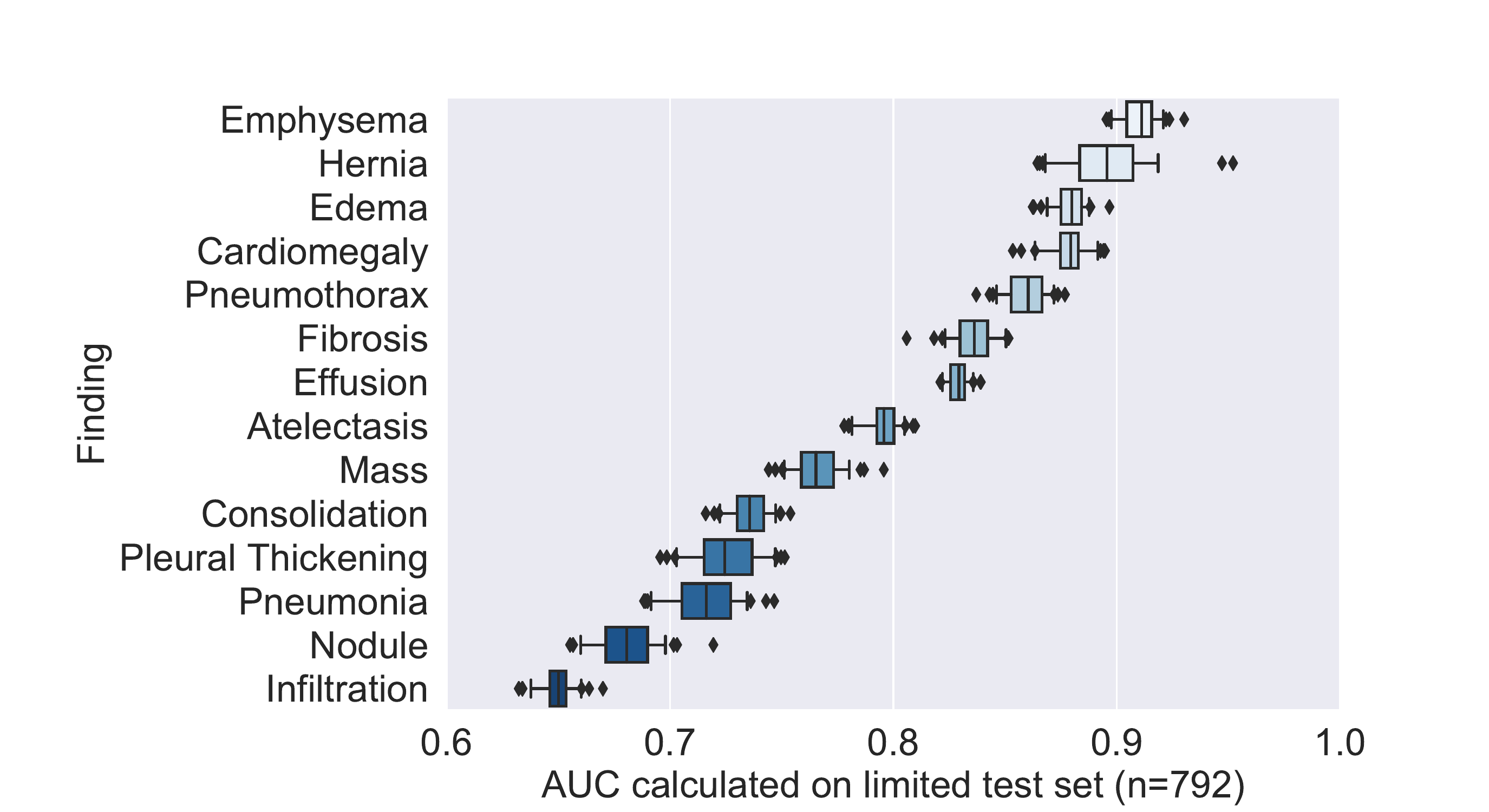}
\label{fig:fig_c}
\end{subfigure}
\caption{Boxplots of AUC across all 50 models in the full test set ($n$=22,433; top panel) and limited test set, representative of test sets used for expert labeling ($n$=792; bottom panel). Despite substantial individual radiograph-level variability, variability in AUC was low for most findings on the full test set due to the large sample size. The limited test set had expectedly wider distributions over AUC.}
\label{fig:auc}
\end{figure}

Averaging model predictions significantly reduced the mean coefficient of variability from 0.543 to 0.169 (t-test 15.96, p-value < 0.0001). The distribution over AUC across models showed a degree of variability in both the full and limited test sets (Figure~\ref{fig:auc}, Table~\ref{table:auc}). In the limited test set, the empirical variability in predictions did not exceed the average DeLong or bootstrap confidence interval for each model (Table~\ref{table:auc}). The DeLong and bootstrap 95\% confidence intervals for AUC contained the mean AUC across models in 99.7\% of cases ($n$=698/700).

\begin{table}[htbp]
\caption{Mean full test set AUC for each finding, variability across repeated model runs, and comparison with DeLong and bootstrap confidence interval for AUC.}
\label{table:auc}
\centering
\begin{tabular}{ccccccc}
\hline
                            & \multicolumn{2}{c}{\begin{tabular}{@{}c@{}}\textbf{Full test set} \\ \textbf{(n=22,433)} \end{tabular}} & \multicolumn{4}{c}{\textbf{Limited test set (n=792)}}                                                                                    \\
\hline
                            
\textbf{Finding}          & 
\begin{tabular}{@{}c@{}}\textbf{Mean} \\ \textbf{AUC}\end{tabular} &
\begin{tabular}{@{}c@{}}\textbf{Empirical} \\ \textbf{95\% CI}\\ \textbf{width}\end{tabular} &
\begin{tabular}{@{}c@{}}\textbf{Mean} \\ \textbf{AUC}\end{tabular} &
\begin{tabular}{@{}c@{}}\textbf{Empirical} \\ \textbf{95\% CI}\\ \textbf{width}\end{tabular} &
\begin{tabular}{@{}c@{}}\textbf{Average} \\ \textbf{DeLong}\\ \textbf{ 95\% CI}\\ \textbf{width}\end{tabular} &
\begin{tabular}{@{}c@{}}\textbf{Average} \\ \textbf{bootstrap}\\ \textbf{95\% CI}\\ \textbf{width}\end{tabular} 
\\
\hline

\textbf{Atelectasis}        & 0.817             & 0.010                             & 0.796             & 0.029                            & 0.077                                 & 0.077                                     \\
\textbf{Cardiomegaly}       & 0.906             & 0.012                             & 0.878             & 0.037                            & 0.083                                 & 0.082                                     \\
\textbf{Consolidation}      & 0.802             & 0.010                             & 0.736             & 0.030                            & 0.097                                 & 0.097                                     \\
\textbf{Edema}              & 0.894             & 0.010                             & 0.879             & 0.025                            & 0.072                                 & 0.071                                     \\
\textbf{Effusion}           & 0.883             & 0.004                             & 0.829             & 0.018                            & 0.065                                 & 0.065                                     \\
\textbf{Emphysema}          & 0.923             & 0.012                             & 0.910             & 0.028                            & 0.067                                 & 0.066                                     \\
\textbf{Nodule}             & 0.772             & 0.011                             & 0.681             & 0.047                            & 0.111                                 & 0.109                                     \\
\textbf{Pneumonia}          & 0.760             & 0.023                             & 0.715             & 0.054                            & 0.137                                 & 0.136                                     \\
\textbf{Fibrosis}           & 0.827             & 0.017                             & 0.836             & 0.033                            & 0.100                                 & 0.100                                     \\
\textbf{Hernia}             & 0.911             & 0.067                             & 0.897             & 0.082                            & 0.108                                 & 0.105                                     \\
\textbf{Infiltration}       & 0.712             & 0.009                             & 0.650             & 0.030                            & 0.087                                 & 0.087                                     \\
\textbf{Mass}               & 0.837             & 0.015                             & 0.766             & 0.040                            & 0.103                                 & 0.103                                     \\
\textbf{Pleural Thickening} & 0.784             & 0.016                             & 0.725             & 0.051                            & 0.112                                 & 0.111                                     \\
\textbf{Pneumothorax}       & 0.873             & 0.012                             & 0.860             & 0.031                            & 0.077                                 & 0.077              \\  \hline
                    
\end{tabular}
\end{table}

\section{Discussion}

\begin{figure*}[h]
  \centering
  \includegraphics[width=1.0\textwidth]{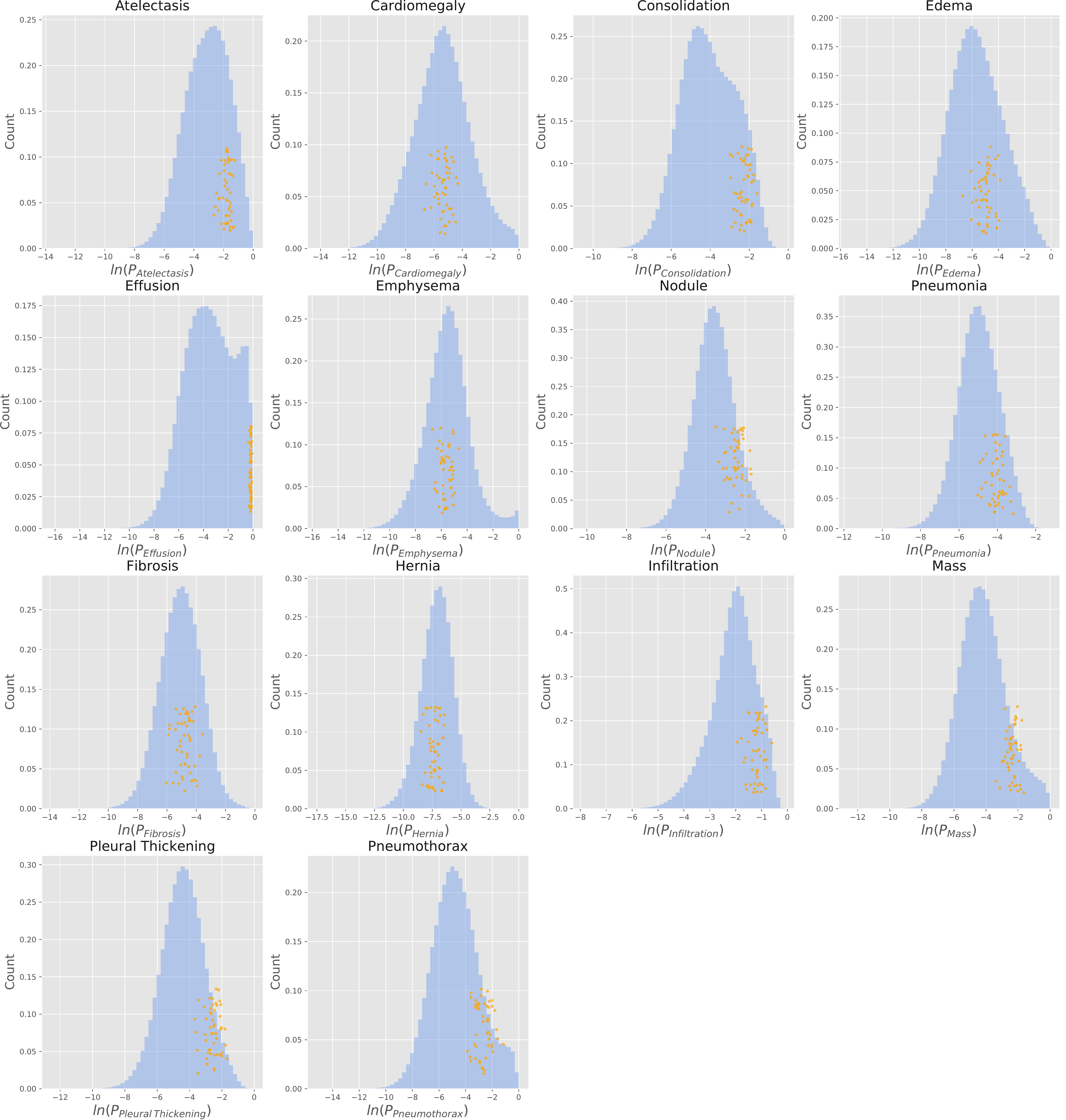}
  \caption{Comparison of the variability in predictions for an example radiograph across models (orange dots; $n$=50 per finding) to distribution of predictions in the full test set (blue histogram; $n$=22,433 cases $\times$ 50 retrainings = 1,121,650 predictions per finding). For many findings, the estimated risk varied substantially (e.g., Pneumonia).}
  \label{fig:lnp_scatter}
\end{figure*}

We found substantial variation among the predicted probability of findings when varying the sampling of batches in the training set (mean coefficient of variation across all findings 0.543, mean $\ln(P_{\text{max}}/P_{\text{min}})$ of 2.45; Figure~\ref{fig:ln-pmax-pmin}, Table~\ref{table-pmax-pmin}). We highlighted a case that demonstrated how predicted probabilities could vary across models (Figure~\ref{fig:lnp_scatter}), shifting its estimated risk relative to the test set population based on the random seed used to train the model. The average case had a 43.0\% percentile range between its highest and lowest estimated probability of disease across all 50 models.

We found that there was variability across models in AUC for all findings. The overall AUC for each finding in the full test set of over 20,000 cases was much more stable than the substantial variability in predictions for individual radiographs. As explained by \citet{delong-article}, \enquote{the area under the population ROC curve represents the probability that, when the variable is observed for a randomly selected individual from the abnormal population and a randomly selected individual from the normal population, the resulting values will be in the correct order (e.g., abnormal value higher than the normal value).} In our case, AUC represents the probability that a randomly selected radiograph that is ground-truth positive for pathology will be assigned a higher score by the CNN than a randomly selected radiograph that is ground-truth negative for pathology. Calculating AUC is thus identical to estimating $p$ for a Bernoulli random variable (i.e., a weighted coin-flip) by repeatedly sampling from this distribution and calculating the average $\hat{p}$ over all draws. As our sample size grows larger and larger, our uncertainty interval over the true value of $p$ (and, equivalently, our uncertainty over AUC) grows progressively narrower. AUC can thus be relatively consistent across CNNs that make variable radiograph-level predictions, provided that these variable predictions are similar overall in their ability to classify positive and negative cases.

The variability in AUC was expectedly wider in the limited test set compared to the full test set. We compared realized variability across models to 95\% confidence intervals estimated by two commonly used methods, DeLong
and bootstrapping, on the limited test set and found that the realized variability did not exceed these estimated bounds. We note that this comparison is limited and not fully powered; we use sample mean instead of unknown population mean, limiting our ability to detect true differences. Nevertheless, it provides evidence that variability in AUC does not grossly exceed the estimates of common statistical tests, and that these tests can be used to compare the performance of different CNNs, provided researchers are aware of their variability. 

Stability in AUC across trained models can mask the wide variation in predictions for a single radiograph, and should not reassure researchers that predictions will remain consistent. In our experiments, each DenseNet-121 \citep{Huang2017-hi} model was initialized with the same pre-trained weights from ImageNet \citep{Russakovsky2015-ib} and trained with the same train/tune/test data, optimizer, and hyperparameters. From this consistent configuration, we fine-tuned each model on the NIH chest radiograph dataset \citep{Wang2017-py}, varying only the order in which training data was batched and presented to the model. The substantial variability we observed in predictions for individual radiographs might have been even wider had we allowed the model's initialization parameters, choice of optimizer, or hyperparameters to vary. \citep{Wilson2017-ca, Choi2019-od}.  \citet{Raghu2019-js} suggested that pre-training may not be necessary to achieve competitive performance on medical imaging tasks.  Our results call to question whether the absence of pre-training may induce additional variability in predictions. 

In the context of healthcare, it is particularly important to remain aware of the variability in individual predictions. If deep learning-based decision support will be deployed in clinical settings, their predictions will alter the diagnoses and treatments given to some patients. Justice, beneficence, and respect for persons are the three ethical principles proposed by the Belmont Report \citep{belmont}, which guides discussion of ethical considerations in medical research. An algorithm that treats identical patients differently challenges the value of justice and potentially leaves the care of patients up to a multi-dimensional coin flip. At the same time, radiologists are also far from perfectly consistent \citep{Bruno2015-cs}. \citet{Rajpurkar2017-nh} observed relatively low inter-rater agreement between the radiologists who contributed the expert labels for pneumonia (0.387 average F1 score comparing each individual radiologist to the majority vote of three other radiologists). Similarly, a study of radiologists at Massachusetts General Hospital found 30\% disagreement between colleagues' interpretations of abdominopelvic CTs and 25\% disagreement for the same radiologist viewing the CT at different times \citep{Abujudeh2010-re, Bruno2015-cs}. Machine learning algorithms may offer an opportunity to improve the consistency of medical decisions, but only if we are attentive to the inconsistency of which they, too, are capable. 



Straightforward workarounds, such as averaging predictions across models \citep{Sollich1996-wj, Lakshminarayanan2017-en}, can substantially mitigate the effect of this individual-level variability (coefficient of variation reduced from 0.543 to 0.169, p-value < 0.0001). Reducing the variability in individual predictions is also likely to improve performance metrics such as AUC; ensembling of CNN predictions has been successfully demonstrated in medical imaging literature \cite{titano, wu, mura, Gulshan2016-mb, Rajpurkar2018-vn, Irvin2019-lc, Pan2019-ux}, primarily to optimize model performance. No prior work to our knowledge examines how variability in the predictions of CNNs for radiologic imaging may translate to the care of individual patients. We encourage researchers to be vigilant of the variability of deep learning models and to provide some measure of how consistently their final (possibly ensembled) model performs in predicting findings for individual patients to assure readers that end users of the model will not be fooled by randomness.

\subsubsection*{Acknowledgments}

We would like to thank the Internal Medicine residency program at California Pacific Medical Center (San Francisco, CA) for giving one of the authors (J.Z.) dedicated time to work on this research while he was a preliminary medicine resident. 

\bibliography{refs}
\bibliographystyle{unsrtnat}

\end{document}